\documentclass{article}

\usepackage[cmex10]{amsmath}
\usepackage{amsfonts}
\usepackage{amssymb}
\usepackage{bbm}

\def\rme{\mathrm{e}}

\usepackage{times}
\usepackage{graphicx} 
\usepackage{subfigure} 

\usepackage{natbib}

\usepackage{algorithm}
\usepackage{algorithmic}

\usepackage{hyperref}

\usepackage[accepted]{icml2016}

\icmltitlerunning{Anomaly Detection and Localisation
                  using Mixed Graphical Models}

\begin{document}
\twocolumn[
\icmltitle{Anomaly Detection and Localisation\\
            using Mixed Graphical Models}

\icmlauthor{Romain Laby}{romain.laby@telecom-paristech.fr}
\icmladdress{CNRS LTCI, T\'el\'ecom ParisTech, Universit\'e Paris-Saclay,
            46 Rue Barrault, 75013 Paris\\
            Thales Airborne Systems, 2 Avenue Gay Lussac, 78990 \'Elancourt}
\icmlauthor{Fran\c{c}ois Roueff}{francois.roueff@telecom-paristech.fr}
\icmladdress{CNRS LTCI, T\'el\'ecom ParisTech, Universit\'e Paris-Saclay,
            46 Rue Barrault, 75013 Paris}
\icmlauthor{Alexandre Gramfort}{alexandre.gramfort@telecom-paristech.fr}
\icmladdress{CNRS LTCI, T\'el\'ecom ParisTech, Universit\'e Paris-Saclay,
            46 Rue Barrault, 75013 Paris}

\icmlkeywords{anomaly detection, mixed graphical models, CUSUM}

\vskip 0.3in
]

\begin{abstract} 
We propose a method that performs anomaly detection and localisation within 
heterogeneous data using a pairwise undirected mixed graphical model.
The data are a mixture of categorical and 
quantitative variables, and the model is learned
over a dataset that is supposed not to contain any anomaly.
We then use the model over temporal data, potentially
a data stream, using a version of the two-sided CUSUM algorithm.
The proposed decision statistic is based on a 
conditional likelihood ratio computed for each variable given the others.
Our results show that this function allows to detect
anomalies variable by variable, and thus to localise
the variables involved in the anomalies more precisely
than univariate methods based on simple marginals.
\end{abstract}

\section{Introduction}

Anomaly detection refers to the task of detecting
anomalous samples within a dataset described by $N$ variables,
also called features.
The localisation is the task that aims at identifying the
subset of variables that are at the origin of the detected anomalies.
While the problem of detection has been extensively studied
in the machine learning literature (see \cite{hodge2004survey}),
the problem of localisation in the presence of dependant
variables remains a challenge.

In this paper, we propose to address this question
using undirected probabilistic graphical models.
Such models are particularly useful to represent the joint distribution over a
set of $N$ random variables $X_1,\dots, X_N$, in an efficient and compact
way. Undirected graphical models are commonly tied to Gaussian random variables
yet recent works have studied the possibility of building models over
heterogeneous variables : \cite{yang2014mixed} proposes a general class of
graphical models where each node-conditional distribution is a member
of a univariate exponential distribution and \cite{lee2015learning,laby:hal-01167391}
investigate the problem of learning
the structure of pairwise graphical model over both discrete and continuous variables.
This is done by optimizing the likelihood or the pseudo-likelihood, penalized with
a Lasso or group Lasso regularisation.

A standard approach to perform online anomaly detection on temporal data such as signals,
is to use the CUSUM algorithm
(see \cite{page1954continuous} and \cite{basseville1993detection}).
In this work we propose a two-sided test with an adapted CUSUM algorithm
to detect anomalies that occur in the conditional distributions
rather than in the marginal. The resulting algorithm allows 
to perform change-point detection and to detect,
variable by variable, continuous or categorical,
the time when the distribution of the data changes
from the ``normal'' distribution.



\section{Mixed Model Presentation}\label{sec:model_presentation}

The definition of graphical models relies on the factorisation of the joint
distribution. 
Pairwise models form a particular class of models where the features are 
grouped in sets of one or two variables. Such models have been widely studied
and have a number of practical advantages \cite{schmidt2010graphical}.
In this paper, we focus on mixed models mixing binary variables 
$X_{\mathcal{C}}=\lbrace X_i,i\in\mathcal{C}\rbrace$ 
(called categorical thereafter), and continuous variables 
$X_{\mathcal{Q}} = \lbrace X_u, u\in\mathcal{Q}\rbrace$ 
(called quantitative thereafter). We have
$X=(X_{\mathcal{C}}, X_{\mathcal{Q}})$,
with values in
$\lbrace 0,1 \rbrace^{\lvert \mathcal{C}\rvert}\times\mathbb{R}^{\lvert \mathcal{Q} \rvert}$. 
We use the pairwise mixed model 
\begin{multline}\label{eq:all}
p_{\Omega}(x) = \frac{1}{Z_{\Omega}}\exp \Biggl(  x_{\mathcal{C}}^T \Theta
x_{\mathcal{C}}\\ 
+ \mu^Tx_{\mathcal{Q}} -\frac{1}{2}x_{\mathcal{Q}}^T\Delta x_{\mathcal{Q}}  
+ x_{\mathcal{C}}^T\Phi x_{\mathcal{Q}} \Biggr) \;,
\end{multline}
where $\Omega=(\Theta,\mu,\Delta,\Phi)$ contains all the parameters of the model.
Here, $\Theta=(\theta_{ij})_{i,j\in\mathcal{C}}$
is a symmetric matrix, $\mu=(\mu_i)_{i\in\mathcal{Q}}\in\mathbb{R}^{\mathcal Q}$,
$\Delta=(\delta_{uv})_{u,v\in\mathcal{Q}}$ is a positive definite symmetric matrix and 
$\Phi=(\phi_{iu})_{i,u\in\mathcal{C}\times\mathcal{Q}}$  is a general matrix.

The model \eqref{eq:all} is a mixture between the classic
Ising Graphical Model (IGM) and Gaussian Graphical Model (GGM).
In the Gaussian model, that is, when $p$ takes 
the form of a Gaussian density, the partition function $Z$ is easy 
to calculate and only requires the calculation of the determinant of a $N\times N$ matrix. 
The Ising model is one of the earliest studied undirected model for modeling 
energy of a physical system involving interactions between atoms (see \cite{ising1925}). 
The Ising model has binary variables, \emph{i.e.} each $x_i$ takes values in $\lbrace -1,1 \rbrace$ 
or $\lbrace 0,1 \rbrace$, depending on the authors. Here we use the state space $\lbrace 0,1 \rbrace$.
The Ising model can be generalized for discrete variables, 
for example with the Potts model \cite{potts53}, but this one can be reparametrized as an IGM using 1-of-$K$ encoding, 
as explained in \cite{bishop06}, \S 4.3.4. In the following,
we will therefore only consider binary categorical variables.

\begin{figure}
\centering
\includegraphics[width=.8\columnwidth]{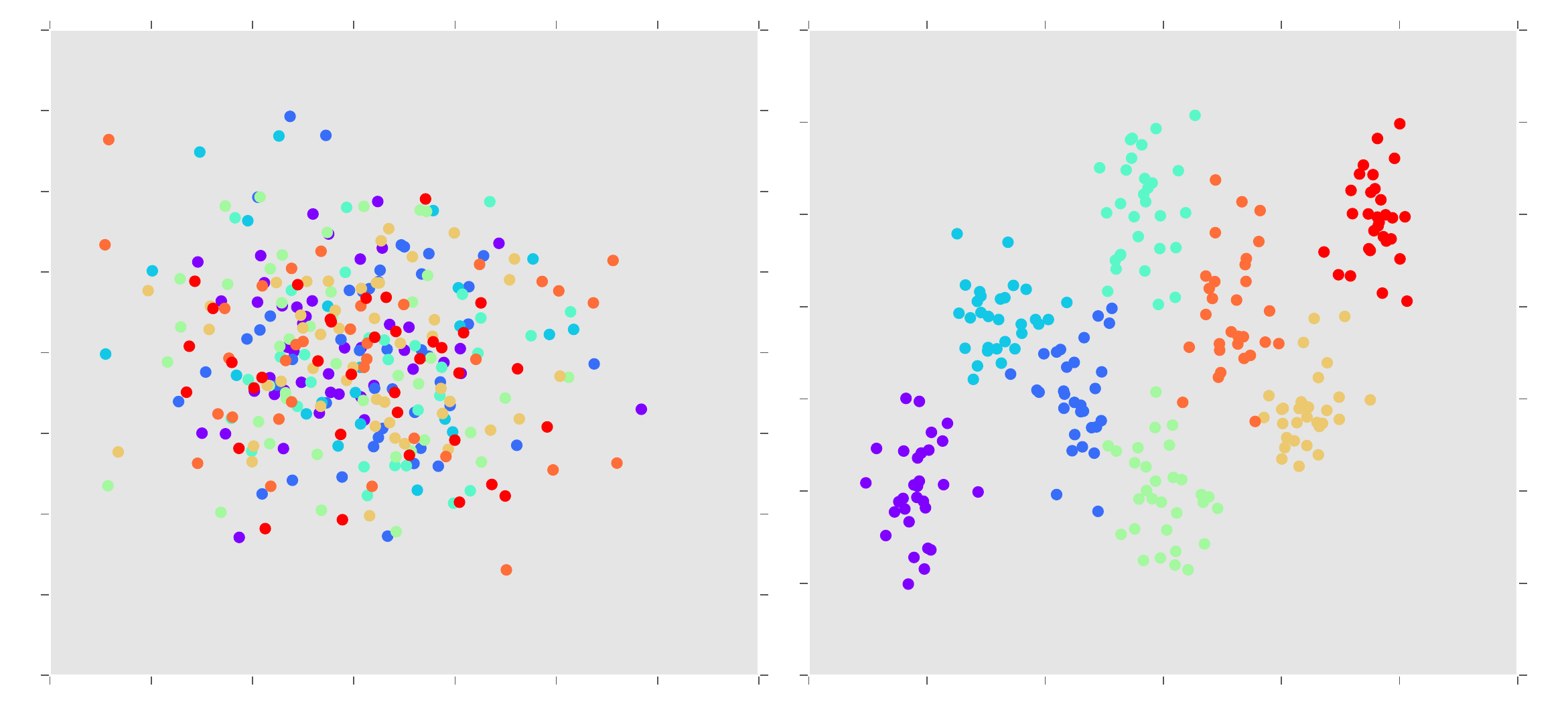}
\caption{I.i.d. samplings of $X_{\mathcal{Q}}$ in two dimensions. The values 
of $X_{\mathcal{C}}$ are represented by $2^3$ different colors. On the left, 
$\Phi = 0$ and on the right, $\Phi \ne 0$.}
\label{fig:mixed_illus}
\end{figure}

To illustrate the model (\ref{eq:all}), we show 
some simulations made with 2 quantitative and 3 categorical variables.
Figure \ref{fig:mixed_illus} shows simulations of $X$
when $\Phi=0$ (the quantitative variables $X_{\mathcal{Q}}$ are independent of the
categorical variables $X_{\mathcal{C}}$ and thus have a Gaussian distribution) and
when $\Phi \ne 0$ ($X_{\mathcal{Q}}$ is not independent of $X_{\mathcal{C}}$ and its
distribution is a mixture of Gaussian distribution). Given $X_{\mathcal{C}}$,
the conditional distribution of $X_{\mathcal{Q}}$ is always Gaussian, namely
\begin{equation}\label{eq:quant_cond_law}
X_{\mathcal{Q}} | X_{\mathcal{C}} \sim \mathcal{N} \left( \Delta^{-1}\left(\mu+ \Phi^T X_{\mathcal{C}}\right), 
\Delta^{-1} \right) \;.
\end{equation}
While, except when $\Phi=0$, the unconditional law of $X_{\mathcal{Q}}$ is not
Gaussian but is a mixture of Gaussian distributions, 
the unconditional law of $X_{\mathcal{C}}$ is again
an Ising model with density
\begin{multline}\label{eq:ising_non_conditional}
p_{\Omega}(x_{\mathcal{C}}) \propto \exp \left(  
x_{\mathcal{C}}^T (\Theta+\Phi\Delta^{-1}\Phi^T/2+ \right.\\
\left.\mathrm{Diag}\left(\Phi\Delta^{-1}\mu\right)) x_{\mathcal{C}}
\right)\;.
\end{multline}
With these two properties, we can design an algorithm to efficiently sample
from the distribution \eqref{eq:all}. Since $p_{\Omega}(X_{\mathcal{Q}},
X_{\mathcal{C}}) = p_{\Omega}(X_{\mathcal{C}})
p_{\Omega}(X_{\mathcal{Q}}|X_{\mathcal{C}})$, one just need to first sample
$X_{\mathcal{C}}$ from \eqref{eq:ising_non_conditional}, using for instance
Wolff's algorithm \cite{wolff1989collective}, and then to sample
$X_{\mathcal{Q}}$ from the conditional Gaussian density
\eqref{eq:quant_cond_law}.  This procedure will be used in our numerical
experiments below.

In this paper, we do not aim at learning a graphical model
but rather at exploiting one for anomaly detection and localisation.
See \cite{yang2014mixed}, \cite{lee2015learning} and \cite{laby:hal-01167391}
for recent works that investigate the task of learning the parameters of a mixed 
undirected graphical model.
%
%
%
%

\section{Anomaly detection and localisation}\label{sec:ano_detection_localisation}

In this section, we present a method to detect and localise
anomalies from a sequence of new data
\mbox{$(X_{\mathcal{C}}^{(t)}, X_{\mathcal{Q}}^{(t)})$},
\mbox{$t=1,2,\dots$}, assuming a reference model $\Omega$
that has already been learned using normal data.

The idea to localise anomalies is to monitor each term of
the log-pseudo-likelihood \cite{besag75} as a function of time.
The CUSUM algorithm \cite{page1954continuous} has been introduced to 
sequentially detect a change in the mean of a random variable.
Since we want to detect an
increase or a decrease, we use the two-sided CUSUM algorithm as proposed in
\cite{basseville1993detection}. For each $t$ and each variable $X_i$, we
define the instantaneous conditional log-likelihood ratio
\begin{equation}\label{eq:likelihood_ratio}
s_i^{(t)}=\log \left( 
\frac{  p \left( X_i^{(t)}|X_{-i}^{(t)} \right)  }  
{p_{\Omega} \left( X_i^{(t)} | X_{-i}^{(t)} \right) } \right)
\end{equation}
where $X_{-i}=\{ X_j, j\in\mathcal{C}\cup\mathcal{Q}, j\ne i\}$,
and a decision statistic defined recursively by $S_i^{(0)}=0$ and
\begin{equation}\label{eq:decision_function}
S_i^{(t)} = \left( S_i^{(t-1)} + s_i^{(t)} \right)^+, ~~t=1, 2, \dots \;,
\end{equation}
where $(z)^+ = \max(z, 0)$. Here $p$ denotes the density of the 
alternative hypothesis, that is, the conditional density of the targeted anomalous
behaviour.

We focus first on the quantitative variables. By \eqref{eq:quant_cond_law},
the conditional distribution of $X_{\mathcal{Q}}^{(t)}$ given $X_{\mathcal{C}}^{(t)}$ 
is the multivariate Gaussian
\mbox{$\mathcal{N}(\nu^{(t)}, \Delta^{-1})$}, with
\mbox{$\nu^{(t)}=\Delta^{-1}(\mu+\Phi^TX_{\mathcal{C}}^{(t)})$}.
It follows that, for all $i\in \mathcal{Q}$, the conditional distribution of
$X_i^{(t)}$ given $X_{-i}^{(t)}$ is Gaussian univariate with mean
$$e_i^{(t)}= \mathbb{E}_{\Omega}[X_i^{(t)}~|~X_{-i}^{(t)}] = \Delta^{-1}_{i, -i} \Delta_{-i, -i} 
\left( X_{\mathcal{Q}-i}^{(t)} - \nu_{-i}^{(t)} \right) + \nu_i^{(t)}$$ and variance 
$$\sigma_i^2= \mathsf{Var}_{\Omega}(X_i^{(t)}~|~X_{-i}^{(t)}) = 
\Delta^{-1}_{ii} - \Delta^{-1}_{i, -i}\Delta_{-i, -i}\Delta^{-1}_{-i, i} \;.$$
We actually see from \eqref{eq:quant_cond_law} that $e_i^{(t)}$ 
depends on $X_c^{(t)}$ and
a fortiori on $t$, whereas it is not the case for $\sigma_i$.

For each quantitative variable $X_i$, $i\in\mathcal{Q}$, we want to detect a change 
in $p_{\Omega}(X_i^{(t)}~|~X_{-i}^{(t)})$.
We define the conditional density $p(X_i^{(t)}~|~X_{-i}^{(t)})$ 
of the alternative hypothesis as a Gaussian density with 
same variance $\sigma_i^2$ and a modified mean
$e_i^{(t)} + \delta\sigma_i$.
The ratio (\ref{eq:likelihood_ratio}) then becomes, for $i \in \mathcal{Q}$,
\begin{align}\label{eq:stat_q}
\nonumber
s_i^{(t)} &= \frac{1}{2} \left( \frac{X_i^{(t)} - e_i^{(t)}}{\sigma_i} \right)^2
- \left( \frac{X_i^{(t)} - (e_i^{(t)} + \delta\sigma_i )}{\sigma_i} \right)^2\\
&= \frac{(X_i^{(t)} - e_i^{(t)})}{\sigma_i} \delta - \frac{1}{2} \delta^2 \;.
\end{align}
Setting $\delta>0$ or $\delta<0$ defines two statistics $S_i^{(t)\uparrow}$
and $S_i^{(t)\downarrow}$ in~(\ref{eq:decision_function}), for detecting respectively
increase and decrease of the conditional mean $e_i^{(t)}$.
In our experiments in Section \ref{sec:applications}, 
we will consider the sum $\bar{S}_i^{(t)}=S_i^{(t)\uparrow} + S_i^{(t)\downarrow}$
in order to detect a change in both possible directions.

Note that, by \eqref{eq:stat_q}, the conditional negative drift under the null
hypothesis (when no changes occur)
of the decision statistic \eqref{eq:decision_function}
is given by \mbox{$\mathbb{E}_{\Omega}[s_i^{(t)}|X_{-i}^{(t)}]=-\delta^2/2$}.

We focus now on the categorical variables. Each variable 
\mbox{$X_i, i\in\mathcal{C}$} has a conditional Bernoulli distribution with mean
\begin{align}
p_i  = \mathbb{E}_{\Omega}[X_i~|~X_{-i}]  = 
 \frac{\rme^{q_{\Omega}(X,i)} }{1+\rme^{q_{\Omega}(X,i)}}\;,
\end{align}
where
\begin{align}
q_{\Omega}(X,i_0)=\theta_{i_0i_0} +
  2\Theta_{i_0,-i_0}X_{-i_0} + \Phi_{i_0,
    \mathcal{Q}}X_{\mathcal{Q}} \;.
\end{align}
In the case of categorical variables, we define the conditional distribution of the
alternative hypothesis as a Bernoulli distribution with mean $a_i^{(t)}$.
The instantaneous log-likelihood ratio is then  given by
\begin{equation}
s_i^{(t)} = X_i^{(t)}\log\frac{a_i^{(t)}}{p_i^{(t)}} + 
(1 - X_i^{(t)})\log \left( \frac{1 - a_i^{(t)}}{1 - p_i^{(t)}} \right) \;.
\end{equation}
We choose $a_i^{(t)}$ such as the drift of the decision 
function \eqref{eq:decision_function} under the null hypothesis is set to the
same value $-\frac{\delta^2}{2}$, as for quantitative variables in \eqref{eq:stat_q}.
This drift is given by computing
$\mathbb{E}_{\Omega}[s_i^{(t)}~|~X_{-i}^{(t)}]$ with $s_i^{(t)}$ as
in~(\ref{eq:stat_q}), yielding the equation
\begin{equation}
p_i^{(t)} \log \frac{a_i^{(t)}}{p_i^{(t)}} + (1 - p_i^{(t)})
\log \left( \frac{1 - a_i^{(t)}}{1 - p_i^{(t)}} \right) = 
-\frac{\delta^2}{2}\;.
\end{equation}
It is easy to show that this equation in $a_i^{(t)}$ (with $\delta$ and $p_i^{(t)}$
fixed) has two distinct solutions 
$a_i^{(t)\uparrow}\in [p_i^{(t)}, 1]$ (associated to the statistic $S_i^{(t)\uparrow}$) and
$a_i^{(t)\downarrow}\in [0, p_i^{(t)}]$ (associated to $S_i^{(t)\downarrow}$),
detecting respectively increase and decrease of the mean $p_i^{(t)}$,
with a conditional negative drift $-\frac{\delta^2}{2}$ under the null hypothesis. For the same reasons 
as with quantitative variables, we will consider
the sum $\bar{S}_i^{(t)}=S_i^{(t)\uparrow} + S_i^{(t)\downarrow}$
in the experiments.

Under the null hypothesis each decision statistic $S_i^{(t)\uparrow}$ or
$S_i^{(t)\downarrow}$ evolves with a negative drift $-\delta^2/2$. Hence, because
of the positive part in~(\ref{eq:decision_function}), it remains close to zero
with high probability. In contrast, under the alternative, the conditional drift
becomes positive and the decision statistic $\bar S_i^{(t)}$ eventually increase above
any arbitrarily high threshold $h$. We thus label as a change time  the first
times $t$ when \mbox{$\bar S_i^{(t)} > h$}. The choice of $\delta$ sets how
sensitive the test is to a close alternative, while the choice of $h$
is a compromise between the false alarm probability over a given horizon and
the delay needed to raise an alarm after a change of distribution. Finally and
most interestingly, the set of indices $i$ for which the alarm is raised 
provides a way to identify the variables for which not only the marginal
distribution has changed but also the conditional one, given all other
available variables. 


\section{Applications on synthetic data}\label{sec:applications}

In this section, we present results of anomaly detection
and localisation with synthetic data.
We suppose here that we have already learned the model parameters $\Omega$
from normal data.
The data are composed of 50 normal observations sampled from the model using the 
algorithm explained at the end of section \ref{sec:model_presentation},
and 50 anomalous observations sampled from an altered model where
one parameter value in $\Omega$ has been modified.

\begin{figure}
\begin{center}
\centerline{\includegraphics[width=.5\columnwidth]{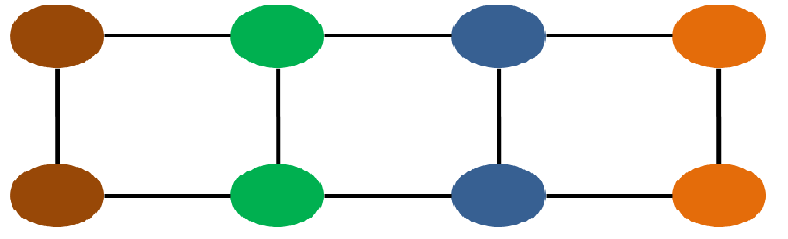}}
\caption{Structure of mixed graphical model used in the experiment. 
The upper and lower layer represents quantitative
and categorical variables. Each $i$-th quantitative and $i$-th
categorical variable have the same color.}\label{fig:hybrid_struct}
\end{center}
\end{figure}

\begin{figure}
\centering
\includegraphics[width=.9\columnwidth, height=3cm]{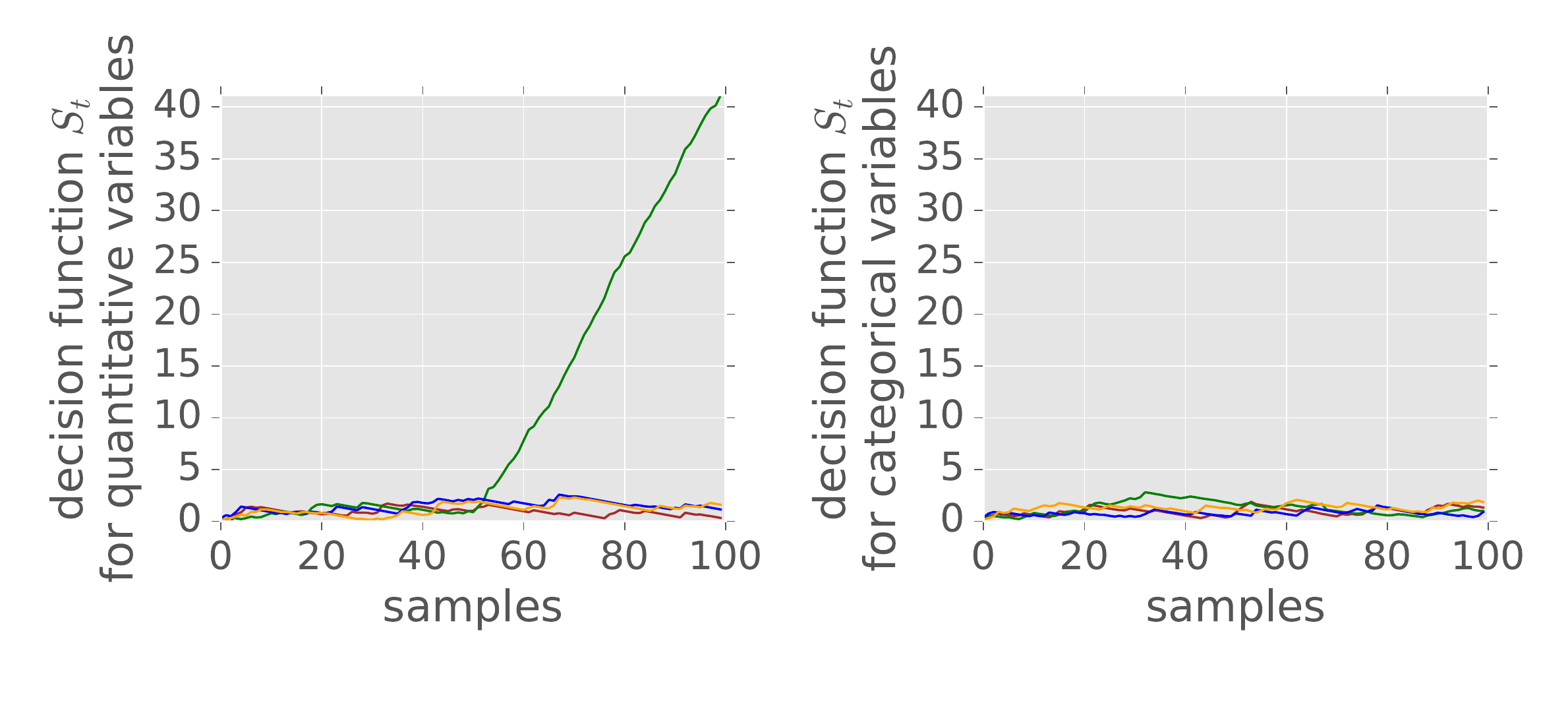}
\includegraphics[width=.9\columnwidth, height=2.5cm]{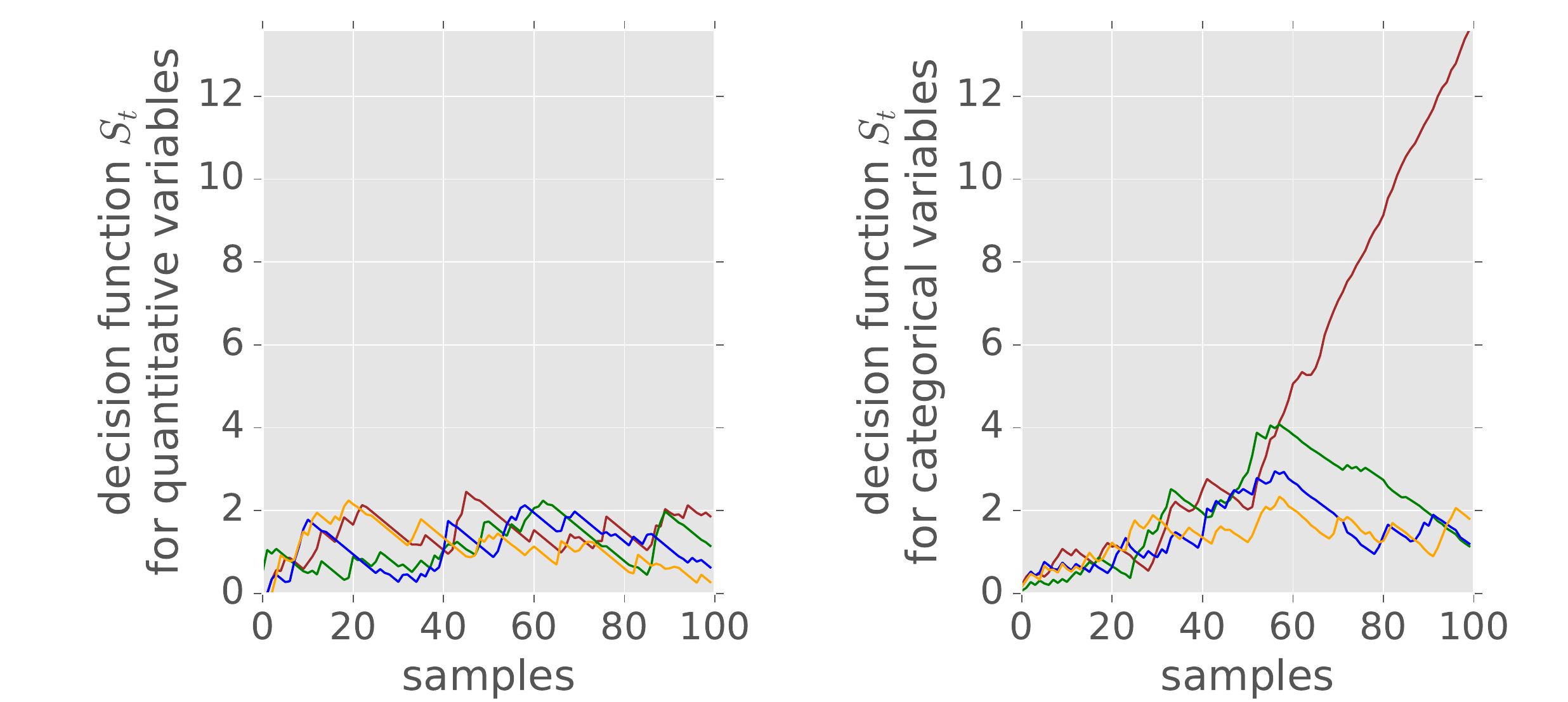}
\includegraphics[width=.9\columnwidth, height=2.7cm]{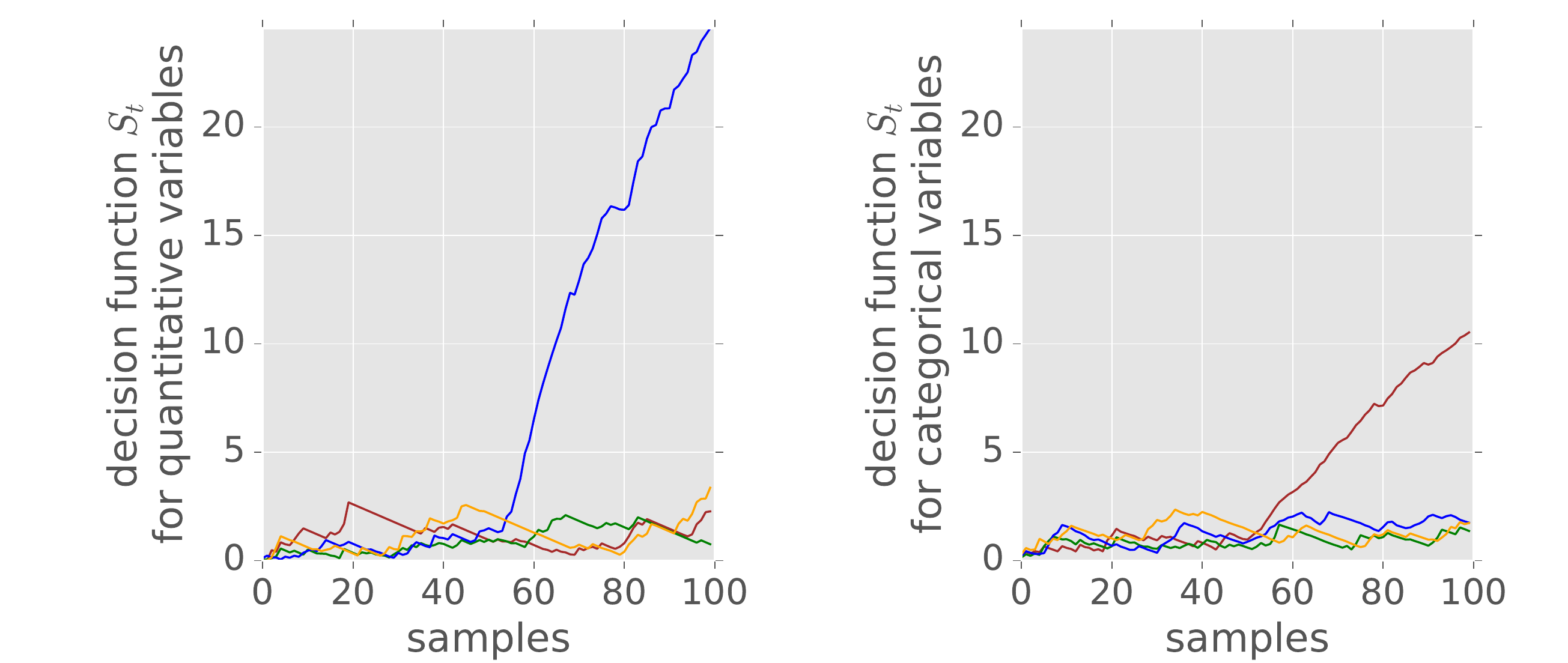}
\caption{Time evolution of $\bar S_i^{(t)}$ for quantitative variables on the left
and categorical variables on the right. The colors of the plots correspond to
the colors of the variables in the graph of Figure~\ref{fig:hybrid_struct}.
Top row~: change on $\mu_1$. Middle
row~: change on $\theta_{0,0}$. Bottom row~: change on $\phi_{0,2}$.
For each experiment, the first 50 samples are sampled with parameter $\Omega$,
and the last 50 samples  are sampled with the modified parameter.}
\label{fig:3_anomalies}
\end{figure}

We use the same model structure as in \cite{lee2015learning}, with 
4 categorical and 4 quantitative variables. 
The model is represented in Fig.~\ref{fig:hybrid_struct}, with a 
colormap that will be kept for all experiments.
The parameters have been chosen as follows:
%
upper and lower diagonal of $\theta$ are filled with .5,
\mbox{$\theta_{i, i} = -\sum_{j\ne i}\theta_{i,j}$} and 0 elsewhere;
$\mu_{i} = 0$, $\Delta_{i,i}=1$, lower and upper diagonal 
of $\Delta$ is filled with .25 and 0 elsewhere;
$\phi_{i,i} = .5$ and 0 elsewhere.
We have tested three different modifications on the parameters of $\Omega$~: 1)
the conditional distribution of the second (green) quantitative variable is
changed by moving $\mu_{1}$ from $0$ to $3$, 2) the conditional distribution of
the first (red) categorical variable is changed by moving $\theta_{0,0}$ from
$-1$ to $-4$ and 3) the conditional distributions of the first (red)
categorical and third (blue) quantitative variable are changed by moving
$\Phi_{0,2}$ from $0.5$ to $2$.  Figure~\ref{fig:3_anomalies} shows the
temporal evolution of the statistic $\bar S_i^{(t)}$ computed for every
variable and for the three kinds of anomalies. As expected, the plots on the
top row show that when changing $\mu_{1}$, only the statistic of the green
quantitative variable is increasing, indicating that the green variable is
carrying alone the change of conditional distribution. The same thing can be concluded for the two others
modifications on $\theta_{0,0}$ and $\phi_{0,2}$.  These results show that our
method correctly detects and localises the changes in the conditional
distributions.

\begin{figure}[h!]
\centering
\includegraphics[width=.7\columnwidth]{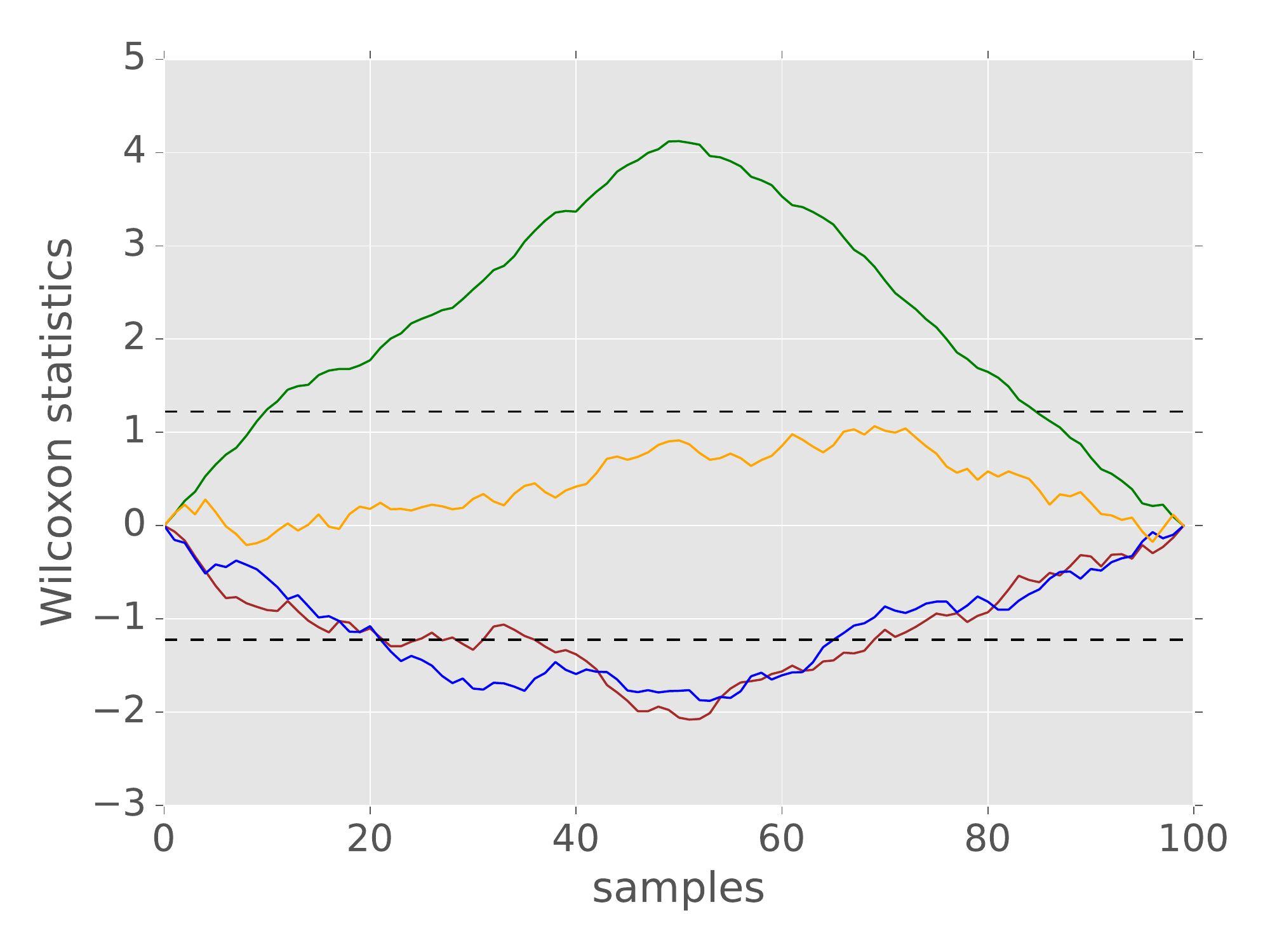}
\caption{Evolution of the Wilcoxon statistic for 100 samples 
of 4 quantitative variables. After the $50^{th}$ sample, 
we have modified $\Omega$ with $\mu_1=3$.
The dashed lines indicate the thresholds for detecting a change with a 5\%
false detection probability.}
\label{fig:wilcoxon}
\end{figure}

We compare our method to the Wilcoxon test presented in 
\cite{lung2011homogeneity}, which is designed to detect changes in 
the distribution of a set of quantitative variables  from batch data.
 In the following, we thus apply this
approach to detect a change of distribution for each quantitative variable.
Figure~\ref{fig:wilcoxon} displays the
statistic of this test as a function of the possible change times. 
When only one change occurs in the data,  this
statistic is expected to approximately have a triangle shape with a maxima
or a minima around the true change time. We use the same dataset as for the experiment
with the anomalies localised on the second (green) quantitative variable, where $\mu_1$
changes from 0 to $3$ at time $t=50$. Figure~\ref{fig:wilcoxon} should thus be
compared with the top row of Figure~\ref{fig:3_anomalies}.
In contrast to online methods such as the one we propose, this Wilcoxon
statistic cannot be computed recursively as it it requires the whole set of
data to be computed. Moreover it is not suited to localise the anomaly since a
change of $\mu_1$, although it only modifies the conditional distribution of
$X_1$ given $X_{-1}$, yields a change of all the marginal distributions. This is why in
Figure~\ref{fig:wilcoxon}, the Wilcoxon statistics display triangle shapes for all the
quantitative variables with a more obvious change for the variables directly
connected to $X_1$.

\section{Conclusion}
In this paper, we proposed an online method that allows to detect anomalies
in a data stream, but more importantly to localise which variables are at the origin
of the problem. By using a mixed undirected graphical model learned
over a set of normal data, we manage to track changes
occurring in the conditional distributions which offers more specific
detections than when studying only marginal distributions.
This method is based on a two-sided CUSUM algorithm, where
decision statistics are computed for every variable and involve
the calculation of conditional likelihoods.

\bibliography{bib}
\bibliographystyle{icml2016}

\end{document}